\documentclass[10pt,twocolumn,letterpaper]{article}

\usepackage{cvpr}              

\usepackage{algorithm}
\usepackage{algorithmic}
\usepackage{float}
\usepackage{enumitem}
\usepackage{amsmath}
\usepackage{amssymb}
\usepackage{booktabs}
\usepackage{multirow}
\usepackage{subcaption}
\usepackage[table]{xcolor}
\usepackage{wrapfig}
\usepackage{pifont}
\usepackage{placeins}
\usepackage{balance}

\definecolor{cvprblue}{rgb}{0.21,0.49,0.74}
\usepackage[pagebackref,breaklinks,colorlinks,allcolors=cvprblue]{hyperref}
\usepackage[accsupp]{axessibility}  

\def\paperID{38492} 
\def\confName{CVPR}
\def\confYear{2026}

\title{Bidirectional Multimodal Prompt Learning with Scale-Aware Training for Few-Shot Multi-Class Anomaly Detection}

\author{Yujin Lee, \: Sewon Kim, \: Daeun Moon, \: Seoyoon Jang, \: Hyunsoo Yoon\thanks{Corresponding Author.}\\
Yonsei University, South Korea\\
{\tt\small \{yjlee9040, sewon3397, dani0403, tjdbs70002, hs.yoon\}@yonsei.ac.kr}
}

\begin{document}
\maketitle

\begin{abstract}
Few-shot multi-class anomaly detection is crucial in real industrial settings, where only a few normal samples are available while numerous object types must be inspected. This setting is challenging as defect patterns vary widely across categories while normal samples remain scarce. Existing vision–language model–based approaches typically depend on class-specific anomaly descriptions or auxiliary modules, limiting both scalability and computational efficiency. In this work, we propose AnoPLe, a lightweight multimodal prompt learning framework that removes reliance on anomaly-type textual descriptions and avoids any external modules. AnoPLe employs bidirectional interactions between textual and visual prompts, allowing class semantics and instance-level cues to refine one another and form class-conditioned representations that capture shared normal patterns across categories. To enhance localization, we design a scale-aware prefix trained on both global and local views, enabling the prompts to capture both global context and fine-grained details. In addition, alignment loss propagates local anomaly evidence to global features, strengthening the consistency between pixel- and image-level predictions. Despite its simplicity, AnoPLe achieves strong performance on MVTec-AD, VisA, and Real-IAD under the few-shot multi-class setting, surpassing prior approaches while remaining efficient and free from expert-crafted anomaly descriptions. Moreover, AnoPLe generalizes well to unseen anomalies and extends effectively to the medical domain.
\end{abstract} 
     
\section{Introduction}
\label{sec:intro}

\begin{figure}
    \centering
    \includegraphics[height=4.05cm]{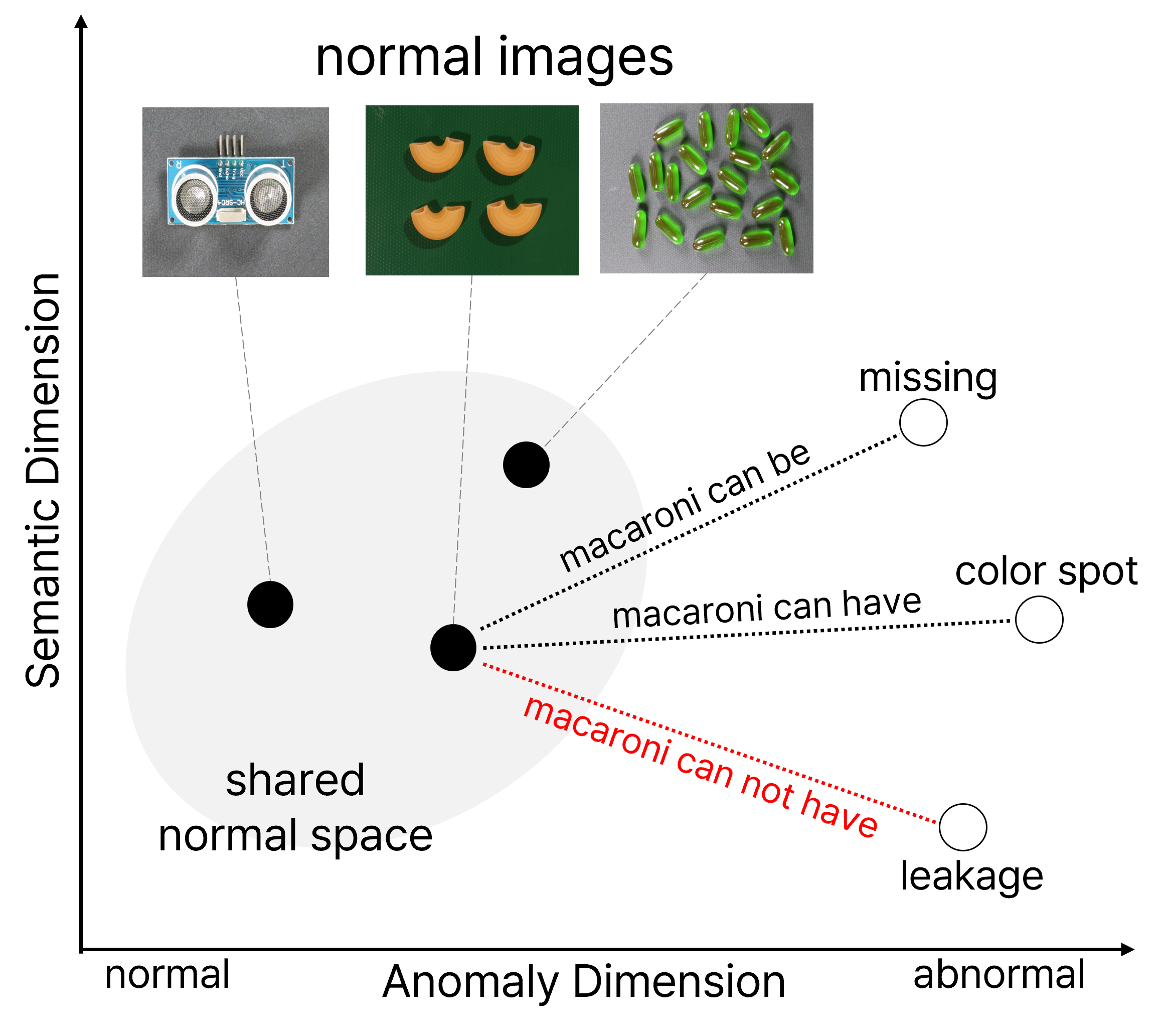}
    \includegraphics[height=3.65cm]{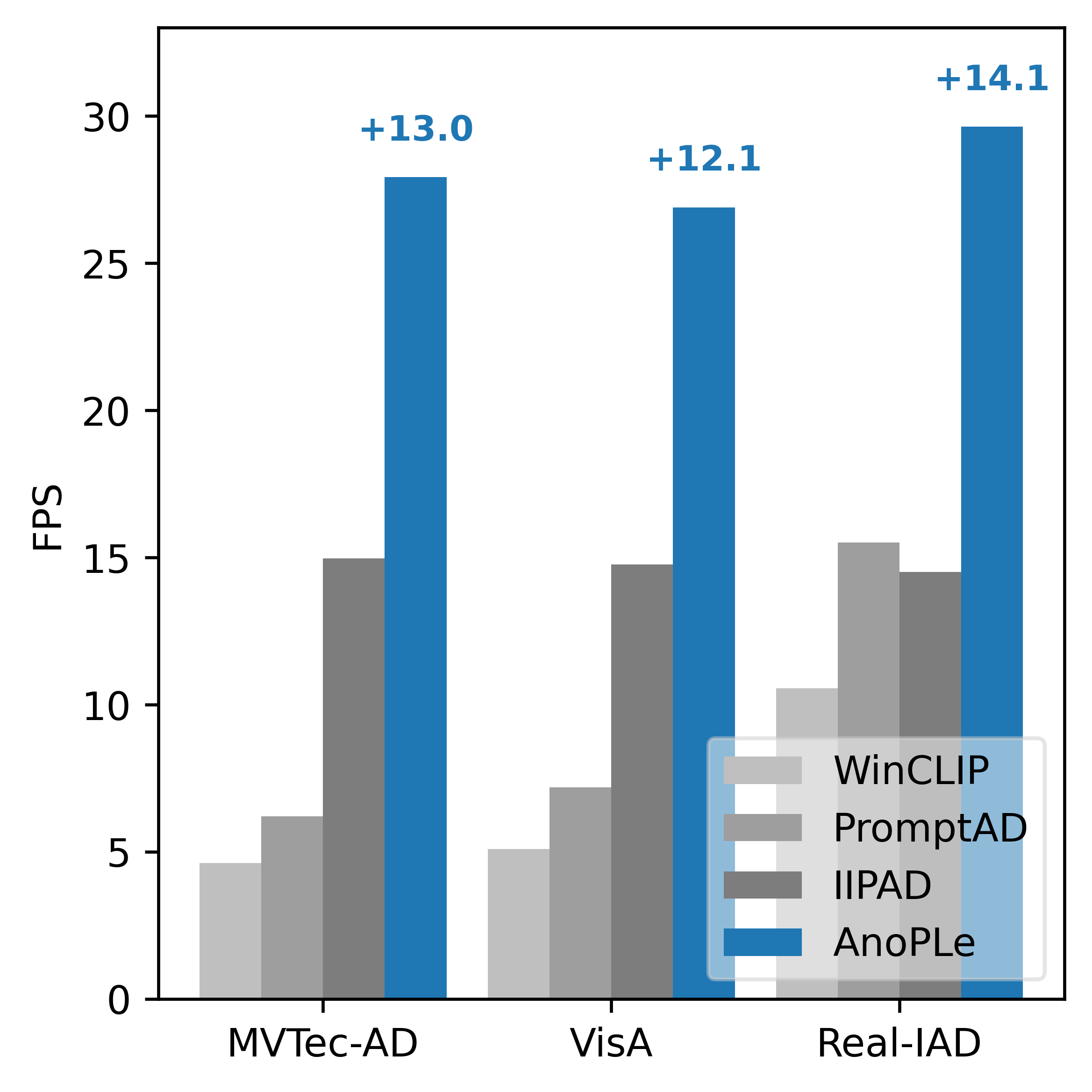}
    \includegraphics[width=0.98\columnwidth]{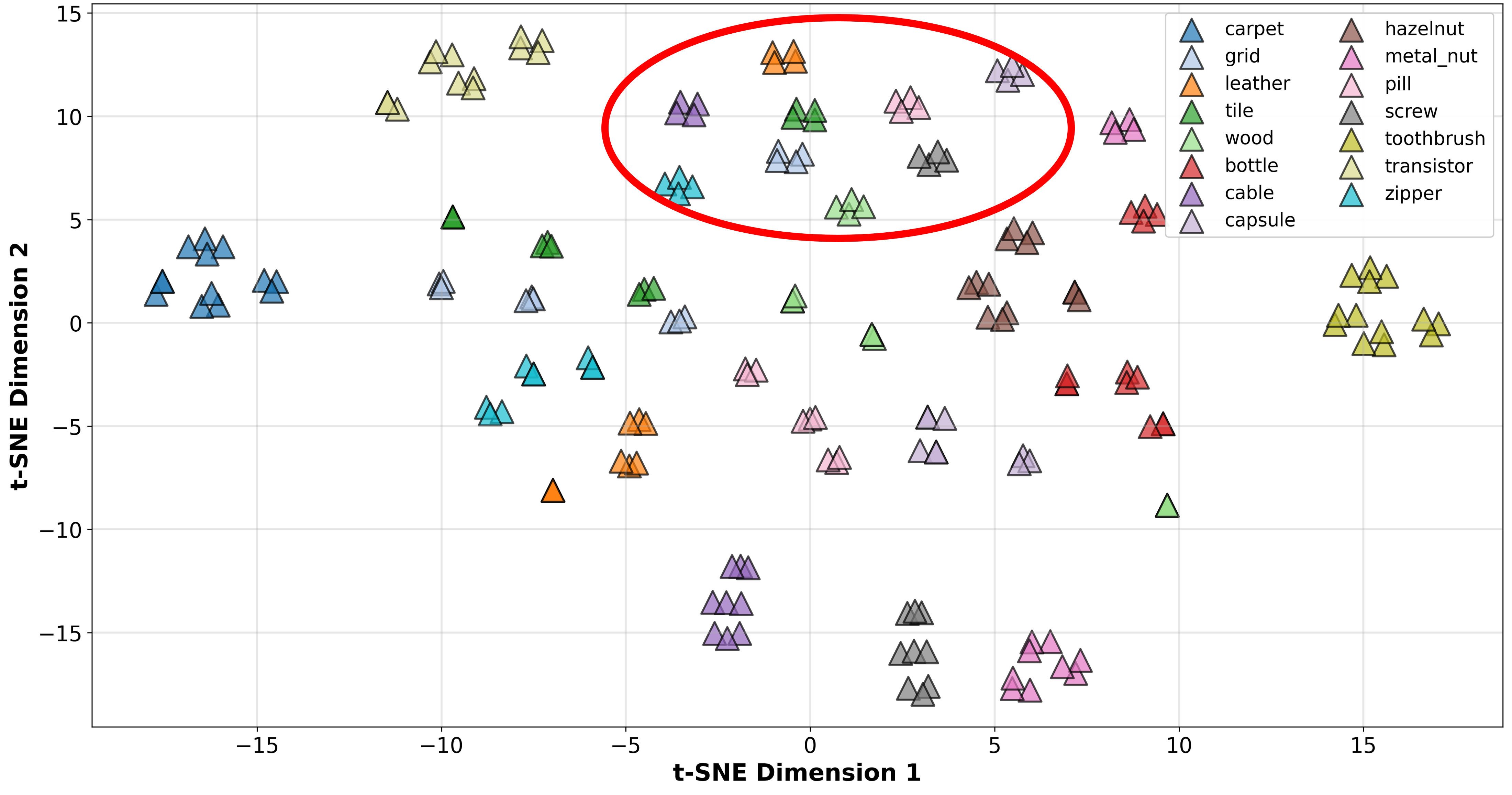}
    \caption{\textbf{(a) Category-dependent anomaly semantics (top-left).} 
Normal samples across categories lie in a shared normal space, while anomalies diverge along category-specific directions. Textual descriptions may align with feasible defects for each category (black), whereas invalid descriptions (red) cause semantic mismatch in text-image alignment. \textbf{(b) Inference speed comparison on MVTec-AD, VisA, and Real-IAD (top-right).} AnoPLe achieves superior efficiency by avoiding the overhead of prior methods, including WinCLIP’s sliding-window inference \cite{jeong2023winclip} and the additional VV-CLIP \cite{li2024promptadcvpr} or Q-Former \cite{Lv2025OneforAllFA} forward passes in PromptAD and IIPAD. This makes AnoPLe faster and more suitable for industrial deployment. \textbf{(c) PromptAD's learned embeddings of MVTec-AD on multi-class setting (bottom).} T-SNE reveals that abnormal features from heterogeneous categories collapse into overlapping regions (circled), indicating semantic confusion despite distinct anomaly descriptions.}
    \label{fig:fps}
\end{figure}

In real-world industrial inspection, the cost of model development, per-category maintenance, and high-quality data collection poses significant practical challenges. Manufacturing environments often face a large number of product categories, frequent category updates, and limited opportunities to curate clean datasets. Motivated by these constraints, the anomaly detection (AD) community has gradually shifted from conventional single-class AD toward more realistic scenarios that better mirror industrial requirements---most notably few-shot and multi-class AD.

Few-shot anomaly detection (FSAD) aims to model normality using only a handful of verified normal samples per category, reducing the need for large curated datasets. With the emergence of vision–language models (VLMs) \cite{radford2021learning,jia2021scaling}, recent FSAD approaches \cite{rudolph2021same,huang2022registration,jeong2023winclip,fang2023fastrecon,li2024promptadcvpr} have leveraged multimodal priors through handcrafted textual templates, or learnable context vectors. While effective, these methods commonly rely on \textit{category-specific} descriptions of defect types, such as “broken fabric’’ or “missing wire.’’ In parallel, multi-class anomaly detection (MCAD) \cite{you2022unified,zhao2023omnial,gao2024learning,he2024diffusion} has emerged as a scalable paradigm that unifies heterogeneous categories within a single model, reducing deployment and maintenance costs. However, it introduces a key challenge: while normal samples differ in appearance across categories, the notion of being “intact” is largely shared, whereas anomalies are diverse and category-dependent. This asymmetry makes it difficult to learn generalizable defect cues without relying on class-specific descriptions. Recent findings \cite{fan2025salvaging} support this view, showing that lightweight class semantics provide a strong inductive bias that organizes heterogeneous categories into a structured representation space. This suggests that category-level cues alone can effectively mitigate inter-class confusion, without handcrafted defect descriptions.

In real industrial environments, two key challenges typically arise together: only a few normal samples are available, and the system must operate across many object categories. This defines the few-shot MCAD setting \cite{huang2022registration, Lv2025OneforAllFA}, where a single model must infer normality from limited observations and still generalize to diverse object types. While the notion of normality (free from breakage, contamination, or geometric irregularity) tends to be consistent across categories, abnormality is highly class-dependent because each category exhibits distinctive defect patterns. To illustrate the inherent limitation of description-based prompts, Figure~\ref{fig:fps}-(a, c) shows how anomaly semantics vary across categories and why using generic textual descriptions often leads to mismatched text–image alignment.

Recent methods often overlook this asymmetry. IIPAD \cite{Lv2025OneforAllFA}, for example, extends single-class AD to the multi-class scenario by collapsing category-specific anomaly descriptions into a unified pool of object-level prompts. This simplification, however, mixes heterogeneous semantics and easily produces confusion, since anomaly characteristics that should remain class-dependent are forced into a shared textual space. IIPAD introduces instance-induced prompts generated by a large Q-Former with auxiliary textual guidance, allowing prompts to adapt dynamically to each sample and partially restoring class-specific sensitivity. Although the concept of instance-induced prompts may mitigate semantic confusion, it inevitably adds substantial computational cost and still depends on descriptions derived from true abnormal samples. Thus, the limitations inherent in description-based designs remain unresolved.

VLMs such as CLIP implicitly encode broad visual concepts of intactness and defectiveness within their latent space, acquired from massive image-text corpora. Instead of manually describing defect types or generating instance-specific prompts, a more scalable direction is to steer these latent priors through multimodal interaction. Motivated by these insights, we propose \textbf{AnoPLe}, a simple and scalable few-shot MCAD framework that avoids handcrafted anomaly descriptions and heavy auxiliary modules. We propose \textbf{bi-directional prompt learning} between textual and visual branches: class names act as lightweight semantic anchors on the textual side, while the visual branch learns instance-level cues directly from image evidence. Leveraging the reciprocal interaction, instance-level prompts refine textual cues with defect-relevant details, and textual cues guide visual features toward class-aware structure---effectively leveraging CLIP's latent normality/abnormality priors without relying on brittle descriptions. 

To enhance fine-grained localization, which is crucial in industrial inspection, AnoPLe uses \textbf{scale-aware prompt learning} that exposes the model to both global context and local details during training. Unlike VV-CLIP-based methods \cite{zhou2023anomalyclip,li2024promptadcvpr,Lv2025OneforAllFA}, our design implicitly encodes multi-scale cues into the prompt without additional inference cost. At test time, only the global prefix is used, maintaining full-image compatibility while retaining strong localization ability. Furthermore, \textbf{alignment loss} propagates pixel-level anomaly cues to the global level, promoting coherent normality modeling even under extreme few-shot constraints. 

Despite requiring no expert-crafted prompts or category-specific tuning, AnoPLe achieves strong performance across diverse industrial benchmarks. Under the one-shot setting, it attains 94.5\%, 86.0\%, and 81.2\% AUROC on MVTec-AD \cite{bergmann2019mvtec}, VisA \cite{zou2022spot}, and Real-IAD \cite{wang2024real}, and accurately localizes anomalies with 90.8\%, 87.5\%, and 88.8\% PRO. The method further generalizes robustly to unseen anomaly types and even medical scenarios, suggesting applicability beyond industrial use cases. In addition, AnoPLe achieves high computational efficiency, running at 27.92, 26.90, and 29.64 FPS on MVTec-AD, VisA, and Real-IAD (Figure~\ref{fig:fps}-(b)). Our main contributions are as follows:
\begin{itemize}
    \item We propose \textbf{AnoPLe}, a unified few-shot MCAD framework that avoids handcrafted anomaly descriptions and heavy external prompt-generation modules.
    \item We introduce \textbf{bi-directional interactions} between textual and visual prompts to learn class-aware yet defect-agnostic representations.
    \item We improve localization through \textbf{scale-aware prefix} and \textbf{alignment loss} that jointly capture global context and fine-grained local cues.
    \item AnoPLe achieves strong few-shot performance even with the single image, showing its practicality and scalability.
\end{itemize}

\section{Related work}

\subsection{Multi-Class Anomaly Detection}

Multi-Class Anomaly Detection (MCAD) aims to unify anomaly detection across multiple object categories within a single model \cite{you2022unified,zhao2023omnial, gao2024learning, he2024diffusion}. Conventional one-class-one-model approaches lack scalability, motivating unified frameworks such as feature registration \cite{huang2022registration}, distribution normalization \cite{guo2024absolute}, adaptive representation learning \cite{you2022unified}, and training-free component-aware matching frameworks \cite{gu2025univad}. A key insight from recent work is the importance of class semantics in improving discrimination across categories. CCL \cite{fan2025salvaging} show that explicitly modeling class-aware structure reduces inter-class confusion and yields more discriminative anomaly features, suggesting that class-level cues provide a strong inductive bias for MCAD. 

Building on this, recent methods incorporate semantic guidance through vision-language models or prompt-based representations, such as AnomalyGPT and IIPAD, which leverage textual or learned semantics to better structure multi-class normality. In contrast, INP-Former \cite{luo2025exploring} adopts an entirely vision-only framework, extracting intrinsic normal prototypes to guide a lightweight reconstruction pipeline. While efficient, relying solely on visual evidence limits the semantic structure available for organizing multi-class representations, which can reduce robustness to unseen categories. Together, these directions highlight two key principles for MCAD: (1) leveraging class semantics to structure multi-class representations and (2) maintaining scalability for real-world industrial settings. However, existing semantic-driven approaches often rely on explicit anomaly descriptions or additional modules, which limits their flexibility in practical few-shot settings.

\subsection{Few-shot Industrial Anomaly Detection}
Few-shot industrial anomaly detection aims to detect anomalies using only limited normal samples. Prior works explore transfer-based methods \cite{huang2022registration} or learning directly from limited normal samples \cite{roth2022towards, li2024promptadcvpr, jeong2023winclip}. With the advent of VLMs such as CLIP \cite{radford2021learning, jia2021scaling}, several studies adapted cross-modal alignment to improve few-shot generalization. WinCLIP \cite{jeong2023winclip} uses handcrafted prompts, while later works introduce learnable adapters \cite{gu2024anomalygpt, zhu2024toward, deng2023anovl} or context vectors \cite{li2024promptadcvpr}. These approaches demonstrate the benefit of textual conditioning but remain limited to single-category settings and often rely on expert-crafted anomaly descriptions, hindering their extension to multi-class scenarios.

\noindent \textbf{Few-shot MCAD.} Combining few-shot industrial anomaly detection with MCAD is challenging  due to the combination of limited normal samples per category and the need to model diverse normal patterns across multiple classes within a unified representation. IIPAD \cite{Lv2025OneforAllFA} provides the first significant attempt in this direction by introducing instance-induced prompts that adapt to individual samples. While effective for category-agnostic anomaly modeling, it relies on a large Q-Former and a pool of expert-defined anomaly descriptions, resulting in high computational cost and limited scalability. Building on recent MCAD findings on the importance of class-aware semantics \cite{fan2025salvaging}, our approach leverages class names as lightweight textual anchors while avoiding anomaly-type-specific prompts.This allows AnoPLe to learn unified, class-sensitive, yet anomaly-type-agnostic representations without relying on expert-crafted anomaly descriptions or heavyweight auxiliary modules.

\section{Method}

\begin{figure*}[ht!]
\begin{center}
    \includegraphics[width=0.99\linewidth]
    {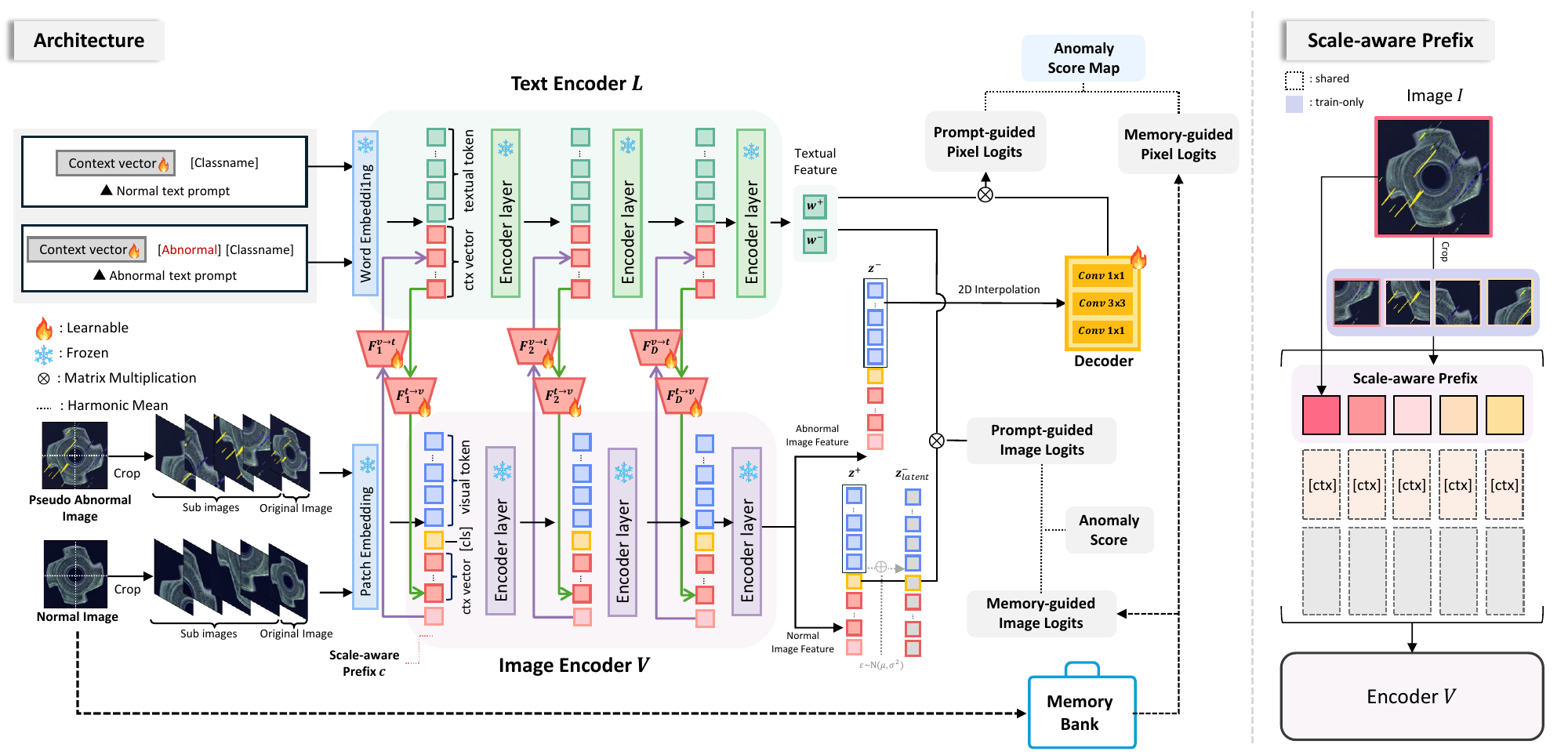}
\end{center}
    \caption{\textbf{Overview of AnoPLe.} AnoPLe employs bidirectional multi-modal deep prompts, establishing strong coupling between textual and visual representations learning from pseudo anomalies and unified text prompts. For localization, we introduce a light-weight decoder to produce pixel logits and train AnoPLe with mixed-scale inputs with a scale-aware prefix. At inference, the anomaly score is computed from our prompt learner and a visual memory bank. \texttt{[ctx]} denotes an abbreviation for the context vector.}
\label{fig:architecture}
\end{figure*}

AnoPLe addresses few-shot MCAD, where only a few normal samples per category are available and anomaly-specific textual descriptions are unavailable. We simulate anomalies in both pixel and latent spaces \citep{zavrtanik2021draem, liu2023simplenet} to learn abnormality cues from normal-only training. AnoPLe then jointly learns textual prompts for class-level semantics and visual prompts for instance-level deviations through bidirectional interactions. This design treats class names as semantic anchors while capturing shared normal patterns across categories without explicit defect descriptions. For localization, we introduce a scale-aware prefix trained on global and local crops and a lightweight decoder for pixel-level prediction, without increasing inference cost. During inference, AnoPLe further leverages a memory bank of normal features for similarity-based anomaly scoring.

\subsection{Multi-Modal Prompts for Anomaly Detection} 
\subsubsection{Textual Prompt Learning}

A central challenge in few-shot MCAD is the semantic heterogeneity of abnormal patterns across categories, which renders explicit defect-level textual descriptions impractical. To establish a minimal yet effective semantic prior, we condition textual prompts solely on class names, omitting a normality token to preserve an intact class prototype. Instead, we introduce a unified abnormality token that encodes a coarse non-normal prior from CLIP pretraining, which is subsequently refined through bidirectional interactions with visual cues during forward propagation to adapt to anomaly detection. \cite{fan2025salvaging} show that such class semantics provide a strong inductive bias that structures categories and mitigates inter-class confusion. Moreover, class names are readily available and consistent with the anomaly detection setting, where supervision is defined by normal–abnormal distinctions rather than category identity.

Guided by these observations, we define the textual inputs for normal and abnormal prompts as
\begin{equation}
    \mathbf{e}_0^{+} = [\texttt{class}], \qquad
    \mathbf{e}_0^{-} = [\texttt{abnormal}]\,[\texttt{class}],
\end{equation}
where \texttt{class} denotes a category name and \texttt{abnormal} is a unified abnormality token shared across all categories.

Learnable textual context vectors $\mathbf{P}_0^{t} \in \mathbb{R}^{C_t \times d_t}$ are prepended to construct the initial text embeddings:
\begin{equation}
    \mathbf{w}_0^{+} = [\mathbf{P}_0^{t},\, \mathbf{e}_0^{+}], \qquad
    \mathbf{w}_0^{-} = [\mathbf{P}_0^{t},\, \mathbf{e}_0^{-}].
\end{equation}

We adopt deep prompting across layers of text encoder $E_T$. At each layer $j$, the contextual vectors are concatenated with the previous hidden state:
\begin{equation}
    \mathbf{w}_j = E_T([\mathbf{P}_j^{t},\, \mathbf{w}_{j-1}]), \qquad j=1,\dots,J.
\end{equation}

\subsubsection{Visual Prompt Learning}

Whereas textual prompts encode class-level priors, the visual branch contributes instance-level cues essential for distinguishing subtle deviations within the same category. An input image $I$ is tokenized into patch embeddings
$\mathbf{z}_0, \mathbf{z}_1, \dots, \mathbf{z}_n$, where $\mathbf{z}_0$ represents the \texttt{[CLS]} token capturing global image information. Learnable visual context vectors $\mathbf{P}_0^{v} \in \mathbb{R}^{C_v \times d_v}$ are prepended to form initial visual normal and abnormal embeddings:
\begin{equation}
    \mathbf{V}_0^{+} = [\mathbf{P}_0^{v},\, \mathbf{z}_0^{+}], \qquad
    \mathbf{V}_0^{-} = [\mathbf{P}_0^{v},\, \mathbf{z}_0^{-}].
\end{equation}

Deep prompting is applied analogously to the textual branch, with $E_V$ denoting the vision encoder:
\begin{equation}
    \mathbf{z}_j = E_V([\mathbf{P}_j^{v},\, \mathbf{z}_{j-1}]), \qquad j=1,\dots,J.
\end{equation}

By operating directly at the instance level, the visual prompts capture fine-grained variations critical for modeling category-specific normality and detecting small defects.

\subsubsection{Bidirectional Coupling of Multi-Modal Prompts}

Since neither explicit anomaly descriptions nor true abnormal images are available, relying on a single modality may yield misaligned or incomplete representations. To enable reciprocal refinement between modalities, we design the textual and visual prompts to interact through learnable linear projections $f_{v\rightarrow t} \in \mathbb{R}^{d_v \times d_t}$ and $f_{t\rightarrow v} \in \mathbb{R}^{d_t \times d_v}$, which project visual prompts into the textual embedding space and vice versa. At each prompting layer $j$, we fuse the prompts from both modalities:
\begin{equation}
    \tilde{\mathbf{P}}_j^{t} = 
    \left[\,\mathbf{P}_j^{t},\; f_{v\rightarrow t}(\mathbf{P}_j^{v})\,\right],
    \;
    \tilde{\mathbf{P}}_j^{v} = 
    \left[\,\mathbf{P}_j^{v},\; f_{t\rightarrow v}(\mathbf{P}_j^{t})\,\right].
\end{equation}

These fused prompts are then injected into the textual and visual encoders:
\begin{equation}
    \mathbf{w}_j = E_T([\tilde{\mathbf{P}}_j^{t}, \mathbf{w}_{j-1}]),
    \qquad
    \mathbf{z}_j = E_V([\tilde{\mathbf{P}}_j^{v}, \mathbf{z}_{j-1}]).
\end{equation}

This coupling allows class-level textual semantics and instance-level cues to iteratively guide one another, yielding representations that are both class-aware and defect-agnostic, aligning with the needs of few-shot MCAD.

\subsection{Bootstrapping Local Semantics from CLIP}\label{sec:scale_method}

\textbf{Scale-Aware Prompt Learning.} CLIP limits its ability to capture local semantics needed for anomaly localization. Rather than modifying CLIP’s architecture or introducing external attention modules, we incorporate scale-awareness directly into the prompt sequence. During training, we present the model with both full-resolution images $I_0$ and non-overlapping sub-image crops $\{I_i \mid 1 \le i \le N\}$ to expose it to global and local semantics. A learnable scale-aware prefix $c \in \mathbb{R}^{(N+1)\times d_v}$ encodes the resolution identity, and the corresponding $c_i$ is prepended to the visual prompt tokens for each input. The resulting behaviors of the global and local prefixes, as well as the $c$, are illustrated in Fig \ref{fig:architecture}. Crucially, this design maintains inference efficiency: only the global scale prefix $c_{N+1}$ is used at test time, avoiding costly multi-scale processing. Embedding scale-awareness directly into the prompts enables CLIP to learn unified representations that respect both global and local cues in a modality-consistent and lightweight manner.

\noindent\textbf{Lightweight Decoder for Local Feature Refinement.} To obtain pixel-level anomaly maps from CLIP's patch embeddings, we employ a lightweight decoder $D$ following \cite{zhou2022extract}. Patch embeddings $\mathbf{z}[1{:}] \in \mathbb{R}^{n \times d_v}$ are upsampled via bilinear interpolation to produce a spatial feature map $\mathbf{F} \in \mathbb{R}^{h \times w \times d_v}$, which is refined by $D$ and projected into the text embedding dimension $d_t$. This refinement introduces local spatial context without increasing inference cost.

\subsection{Loss Function}

\subsubsection{Pixel-Level Loss}

Pixel-level predictions $\hat{\mathbf{M}} = \langle D(\mathbf{z}), \mathbf{w} \rangle$ are supervised using simulated ground-truth masks $\mathbf{M}$. The loss is
\begin{equation}
\mathcal{L}_{pixel} = \mathcal{L}_{dice}(\hat{\mathbf{M}}, \mathbf{M}) + \mathcal{L}_{focal}(\hat{\mathbf{M}}, \mathbf{M}),
\end{equation}
where $\mathcal{L}_{dice}$ encourages overlap with anomalous regions and $\mathcal{L}_{focal}$ accounts for the rarity of anomalous pixels.

\subsubsection{Image-Level Loss}

We compute image-level normality scores using a contrastive cross-entropy loss:
\begin{equation}
\hat{p} = \frac{\exp(\langle \mathbf{z}^{i}_0, \mathbf{w}^i \rangle / \tau)}{\sum_{i\in\{-,+\}}\exp(\langle \mathbf{z}^{i}_0, \mathbf{w}^i \rangle / \tau)},
\end{equation}
where $i\in\{+,-\}$ denotes normal and abnormal text prompts. Text embeddings are averaged across all classes following \cite{jeong2023winclip}. We include pixel-space perturbations ($\mathbf{z}^-$) and latent-space perturbations ($\mathbf{z}^-_{\mathrm{latent}} = \mathbf{z}^+ + \epsilon$ with $\epsilon \sim N(\mu,\sigma^2)$), yielding
\begin{equation}
\mathcal{L}_{img} = \mathcal{L}_{ce}(\mathbf{z}^+, \mathbf{z}^-, \mathbf{w})
+ \mathcal{L}_{ce}(\mathbf{z}^+, \mathbf{z}^-_{\mathrm{latent}}, \mathbf{w}).
\end{equation}

\subsubsection{Alignment Between Local and Global Semantics}\label{sec:align_loss}

To improve consistency between pixel-level and global representations, we introduce $\mathcal{L}_{align}$. Pixel logits $\hat{\mathbf{M}}$ weight the decoder outputs to compute
\begin{equation}
\mathbf{s} = \sum_{(i,j)} \hat{\mathbf{M}}_{ij} \circ D_{ij}(\mathbf{z}),
\end{equation}
which aggregates local evidence aligned with anomalous regions. Alignment is enforced by cosine similarity:
\begin{equation}
\mathcal{L}_{align} = 1 - \langle \mathbf{z}_0, \mathbf{s} \rangle.
\end{equation}
The final training objective is
\begin{equation}
\mathcal{L} = \mathcal{L}_{pixel} + \mathcal{L}_{img} + \mathcal{L}_{align}.
\end{equation}

\subsection{Anomaly Score}

Following \cite{jeong2023winclip, li2024promptadcvpr}, we compute memory-assisted anomaly maps by comparing patch features $\mathbf{F}$ from the $i$-th layer of $E_V$ with a visual memory bank $\mathbf{R}$:
\begin{equation}
   \mathbf{M}_{mem}^{ij} = \min_{\mathbf{r}\in\mathbf{R}} \frac{1}{2}(1 - \langle \mathbf{F}_{ij}, \mathbf{r} \rangle).
\end{equation}

We fuse $\mathbf{M}_{mem}$ and the decoder output $\hat{\mathbf{M}}_q$ using a harmonic mean:
\begin{equation}
    \mathbf{M}_q = \frac{1}{(1/\hat{\mathbf{M}}_q) + (1/\mathbf{M}_{mem})}.
\end{equation}

The final anomaly score aggregates image-level and pixel-level signals:
\begin{equation}
    \mathrm{score} = \frac{1}{(1/\hat{p}) + (1/\max \mathbf{M}_q)}.
\end{equation}

\begin{table*}[h!]
\centering
\caption{\textbf{Anomaly detection and localization performance on the MVTec, VisA, and Real-IAD benchmarks.} 
For MVTec-AD and VisA, we reproduce each baseline under our unified training configuration. For Real-IAD, we adopt the reported values from \cite{luo2025exploring}. The highest values are highlighted in bold, while the runner-up methods are underlined.}
\resizebox{\linewidth}{!}{%
\begin{tabular}{@{}cccccccccccccccc@{}}
\toprule
\multirow{3}{*}{Setup}  & \multirow{3}{*}{Method} 
& \multicolumn{4}{c}{MVTec} & \multicolumn{4}{c}{VisA} & \multicolumn{4}{c}{Real-IAD} \\
\cmidrule(lr){3-6} \cmidrule(lr){7-10} \cmidrule(lr){11-14}
&& \multicolumn{2}{c}{Image-level} & \multicolumn{2}{c}{Pixel-level} 
& \multicolumn{2}{c}{Image-level} & \multicolumn{2}{c}{Pixel-level}
& \multicolumn{2}{c}{Image-level} & \multicolumn{2}{c}{Pixel-level} \\
&& AUROC & AUPR & AUROC & PRO & AUROC & AUPR & AUROC & PRO & AUROC & AUPR & AUROC & PRO \\ \midrule
\multirow{8}{*}{1-shot} 
& SPADE \cite{cohen2020sub} & 73.5 & 87.5 & 59.2 & 27.3 & 68.2 & 72.3 & 61.5 & 37.6
& 51.2 & 45.6 & 59.5 & 19.3 \\
& PatchCore \cite{roth2022towards} & 66.5& 80.6& 88.4 & 66.9 
& 69.8 & 70.8 & 88.8 & 70.0
& 59.3 & 55.8 & 89.6 & 60.5 \\
& WinCLIP \cite{jeong2023winclip} & 77.5 & 87.2 & 86.0 & 70.8
& 70.0 & 73.1 & 84.9 & 61.2 
& 69.4 & 56.8 & 91.9 & 71.0 \\
& PromptAD \cite{li2024promptadcvpr} & 91.2  & 95.4& 93.5 & 86.1 
& 82.4 & 86.0 & 95.1 & 77.8
& 52.2 & 41.6 & 84.9 & 58.4 \\
& IIPAD \cite{Lv2025OneforAllFA} & 94.2 & \underline{97.2} & \underline{96.4} & 89.8
& \underline{85.4} & \underline{87.5} & \underline{96.9} & \underline{87.3}
& -- & -- & -- & -- \\
& INP-Former \cite{luo2025exploring} & \textbf{94.7} & 97.1 & \textbf{96.7} & \underline{90.7}
& {84.0} & {85.9} & {96.7} & {84.0}
& \textbf{84.4} & \textbf{81.9} & \textbf{97.9} & \textbf{88.8} \\
& \textbf{AnoPLe} & \underline{94.5} & \textbf{97.4} & {96.0} & \textbf{90.8}
& \textbf{86.0} & \textbf{87.7} & \textbf{97.0} & \textbf{87.5}
& \underline{81.2} & \underline{80.1} & \underline{96.7} & \underline{88.0} \\
\midrule
\multirow{8}{*}{2-shot} 
& SPADE \cite{cohen2020sub} & 75.5 & 88.4  & 59.2 & 27.1
& 69.5  & 74.9  & 61.5 & 37.5 
& 50.9 & 45.5 & 59.5 & 19.2 \\
& PatchCore \cite{roth2022towards} & 72.7 & 84.8& 90.6& 72.5
& 74.0& 72.7 & 90.1& 72.1
& 63.3 & 59.7 & 92.0 & 66.1 \\
& WinCLIP \cite{jeong2023winclip} & 78.9 & 88.4 & 86.2 & 71.4 
& 70.3 & 73.5 & 85.3 & 62.0
& \underline{70.9} & 58.7 & 93.2 & 74.7 \\
& PromptAD \cite{li2024promptadcvpr} & 93.4& 96.5& 94.0 & 86.7
& 81.4  & 85.9& 95.1  & 75.4
& 57.7 & 41.1 & 86.4 & 61.0 \\
& IIPAD \cite{Lv2025OneforAllFA} & {95.7} & {97.9} & \underline{96.7} & {90.3}
& {86.7} & \underline{88.6} & {97.2} & {87.9}
& -- & -- & -- & -- \\
& INP-Former \cite{luo2025exploring} & \textbf{96.4} & \textbf{98.1} & \textbf{97.2} & \textbf{91.9}
& \textbf{88.5} & \textbf{90.6} & \textbf{97.6} & \underline{87.4}
& \textbf{86.4} & \textbf{84.1} & \textbf{98.2} & \textbf{90.5} \\
& \textbf{AnoPLe} & \underline{95.8} & \underline{98.0} & {96.3} & \underline{91.4}
& \underline{86.9} & \underline{88.6} & \underline{97.3} & \textbf{88.1}
& \underline{82.8} & \underline{82.2} & \underline{97.4} & \underline{89.9} \\
\midrule
\multirow{8}{*}{4-shot} 
& SPADE \cite{cohen2020sub} & 77.4 & 89.1 & 59.2 & 27.2
& 70.5 & 73.5& 61.5& 37.5
& 50.8 & 45.8 & 59.5 & 19.2 \\
& PatchCore \cite{roth2022towards} & 76.8& 87.8 & 92.1& 75.8
& 79.3 & 81.8 & 93.0& 80.0
& 66.0 & 62.2 & 92.9 & 68.6 \\
& WinCLIP \cite{jeong2023winclip} & 78.7& 88.0& 86.4 & 71.9
& 70.5& 73.4 & 85.6 & 62.0
& 73.0 & 61.8 & 93.8 & 76.4 \\
& PromptAD \cite{li2024promptadcvpr} & 94.0& 96.9 & 94.6 & 88.1
& 85.9 & 89.1 & 95.8 & 78.2 
& 59.7 & 43.5 & 86.9 & 61.9 \\
& IIPAD \cite{Lv2025OneforAllFA} & {96.1} & \underline{98.1} & \textbf{97.0} & {91.2}
& \underline{88.3} & \underline{89.6} & {97.4} & \underline{88.3}
& -- & -- & -- & -- \\
& INP-Former \cite{luo2025exploring} & \textbf{96.6} & \underline{98.1} & \textbf{97.0} & \textbf{91.8}
& \textbf{89.3} & \textbf{90.5} & \textbf{98.0} & {88.1}
& \textbf{88.3} & \textbf{85.6} & \textbf{98.4} & \textbf{91.0} \\
& \textbf{AnoPLe} & \underline{96.4} & \textbf{98.4} & \underline{96.5} & \underline{91.3}
& {87.5} & {89.2} & \underline{97.5} & \textbf{88.5}
& \underline{83.2} & \underline{81.5} & \underline{97.4} & \underline{89.5} \\
\bottomrule
\end{tabular}%
}
\label{tab:main_res}
\end{table*}

\section{Experiments}

\subsection{Experimental Details}
We conduct experiments on three widely used industrial anomaly detection benchmarks: MVTec-AD \cite{bergmann2019mvtec}, VisA \cite{zou2022spot}, and Real-IAD \cite{wang2024real}. Following standard practice \cite{bergmann2019mvtec, roth2022towards}, we report image-level AUROC/AUPR and pixel-level AUROC/PRO to evaluate both detection and localization performance. AnoPLe is built upon the CLIP ViT-B/16+ backbone, consistent with recent VLM-based approaches \cite{jeong2023winclip, li2024promptadcvpr}. For multi-scale training, input images are resized to 480×480 and split into four non-overlapping crops, while the original full-resolution view is resized to 240×240. These global and local views are used to construct the scale-aware prompting scheme. Additional hyperparameters and training details are provided in the supplementary materials.

\subsection{Main Results}



In Table \ref{tab:main_res}, across all benchmarks, AnoPLe delivers strong performance in both image- and pixel-level metrics under the few-shot multi-class setting. Classical coreset-based methods such as SPADE \cite{cohen2020sub} and PatchCore \cite{roth2022towards} degrade substantially as category diversity increases (poorly performing in Real-IAD). Recent prompt-based methods \cite{jeong2023winclip, li2024promptadcvpr} still suffer from  semantic confusion when handcrafted anomaly descriptions are collapsed into a unified multi-class space. As particularly visualized in Figure~\ref{fig:fps}-(c), PromptAD’s abnormal embeddings from different categories often collapse into overlapping clusters, indicating semantic confusion when description-based prompts are forced into a unified multi-class space. In contrast, the proposed bidirectional prompt learning enables us to extract rich multi-modal cues to be shared across categories without relying on defect-type descriptions, producing consistently robust results across datasets.

\begin{table}[t]
\centering
\caption{\textbf{Results on unseen classes.} ``Train” denotes the average performance on classes included during training, and ``Held-out” denotes the performance on the excluded unseen class. I-AUC and P-AUC refer to Image AUROC and Pixel AUROC, respectively.}
\label{tab:class_holdout}
\resizebox{\columnwidth}{!}{
\begin{tabular}{llcccc}
\toprule
\multirow{2}{*}{\textbf{Dataset}} & \multirow{2}{*}{\textbf{Class}} 
& \multicolumn{2}{c}{\textbf{I-AUC}} & \multicolumn{2}{c}{\textbf{P-AUC}} \\
\cmidrule(lr){3-4} \cmidrule(lr){5-6}
 &  & INP-F & AnoPLe & INP-F & AnoPLe \\
\midrule
\multirow{2}{*}{\textbf{MVTec}} 
& Train & 93.9 & 93.8 & 96.7 & 95.0 \\
& Held-out & 67.7 (\textbf{\color{red}{26.2}}) & \textbf{87.5} (\textbf{\color{blue}{6.3}}) 
& 81.0 (\textbf{\color{red}{15.7}}) & \textbf{89.5} (\textbf{\color{blue}{5.5}}) \\
\midrule
\multirow{2}{*}{\textbf{Visa}} 
& Train & 87.0 & 83.9 & 97.1 & 95.8 \\
& Held-out & 58.7 (\textbf{\color{red}{28.3}}) & \textbf{72.8} (\textbf{\color{blue}{11.1}})
& 89.7 (\textbf{\color{red}{7.4}}) & \textbf{91.7} (\textbf{\color{blue}{4.1}}) \\
\midrule
\multirow{2}{*}{\textbf{RealIAD}} 
& Train & 83.9 & 79.3 & 97.8 & 96.5 \\
& Held-out & \textbf{63.1} (\textbf{\color{red}{20.8}}) & 58.5  (\textbf{\color{blue}{20.8}})
& 91.3  (\textbf{\color{red}{6.5}}) & \textbf{93.5}  (\textbf{\color{blue}{3.0}})\\
\bottomrule
\end{tabular}
}
\end{table}

We further compare AnoPLe with INP-Former \cite{luo2025exploring}, a vision-only MCAD framework. Although INP-Former reports slightly higher values on a few metrics (e.g., +0.2 AUROC on MVTec image-level under 1-shot and +0.2–0.3 on VisA image-level across settings), the differences remain within a marginal range of 0.1–1.0 across all benchmarks. More importantly, its lack of textual grounding makes it less stable as category variation increases. AnoPLe instead leverages class-name prompts and multimodal interactions to maintain competitive performance across all datasets while offering significantly improved robustness to unseen categories and unseen anomaly types (Table \ref{tab:class_holdout}). This demonstrates that multimodal prompting provides a more scalable and reliable foundation for real-world few-shot MCAD settings.

\subsection{Results in Unseen Anomaly Scenarios}


Real-world industrial environments frequently expose models to unseen object categories or previously unobserved anomaly types. Methods that depend on detailed defect-specific textual descriptions, such as PromptAD, tend to degrade under such conditions. AnoPLe alleviates this issue by using class names as textual anchors while avoiding dependence on anomaly-type descriptions. This design encourages the formation of category-aware yet defect-agnostic representations. In contrast, INP-Former operates purely on visual features without textual grounding, making it less flexible when encountering new categories.

To evaluate these generalization properties, we perform two complementary experiments.
Table \ref{tab:class_holdout} presents a leave-one-class-out analysis, where a category is excluded during training. AnoPLe shows substantially smaller performance degradation on the unseen class than INP-Former, indicating stronger category-level generalization. Table \ref{tab:unseen_anomalies} examines robustness to unseen anomaly types. Removing defect-type descriptions (e.g. ``manipulated front," ``with contamination," ``missing," and ``with extra") from PromptAD to construct PromptAD* leads to large drops in I-AUROC, confirming its reliance on detailed textual priors. AnoPLe, which does not require such descriptions, remains stable even when encountering anomaly types never referenced during training. These results demonstrate that AnoPLe provides strong robustness to both unseen categories and unseen anomaly types, reinforcing its practicality for real-world industrial anomaly detection where new objects and novel defect modes routinely emerge.


\begin{table}[]
    \centering
    \caption{\textbf{Results on unseen anomalies for selected categories from MVTec-AD and VisA, respectively.} PromptAD* excludes anomaly descriptions from the prompt pool of PromptAD.}
    \resizebox{\linewidth}{!}{
    \begin{tabular}{lcccc}
        \toprule
        \textbf{Method} & \multicolumn{2}{c}{\textbf{MVTec-AD}} & \multicolumn{2}{c}{\textbf{VisA}} \\
        \cmidrule(lr){2-3} \cmidrule(lr){4-5}
        & \textbf{screw} & \textbf{bottle} & \textbf{macaroni2} & \textbf{pcb4} \\
        \midrule
        PromptAD & 55.7 & 96.9 & 80.0 &  94.6 \\
        PromptAD* & 44.7 (\textbf{\color{red}{10.0}}) & 93.8 (\textbf{\color{red}{3.1}}) & 73.7 (\textbf{\color{red}{6.3}}) & 90.4 (\textbf{\color{red}{4.2}}) \\ \midrule
        \textbf{AnoPLe (Ours)} & 53.9 & 95.1 & 78.5 & 91.2 \\
        \bottomrule
    \end{tabular}
    }
    \label{tab:unseen_anomalies}
\end{table}

\subsection{Qualitative Results}


\begin{figure}[]
\centering
\includegraphics[width=\columnwidth]{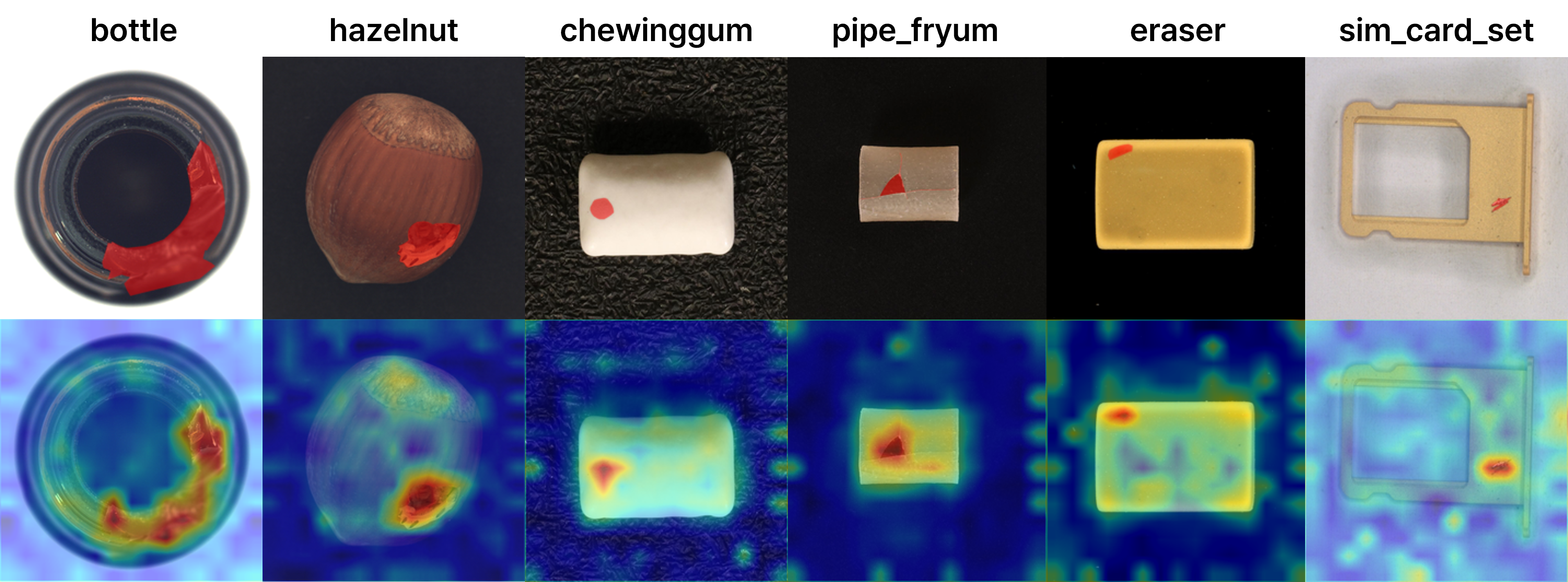}
\caption{\textbf{Qualitative results on MVTec-AD, VisA, and Real-IAD for 1-shot pixel-level anomaly detection.} The top row shows ground-truth anomaly masks overlaid on the input images, while the bottom row displays the anomaly maps from AnoPLe.}
\label{fig:qualitative}
\end{figure}

\begin{figure}[]
    \centering
    \includegraphics[width=\columnwidth]{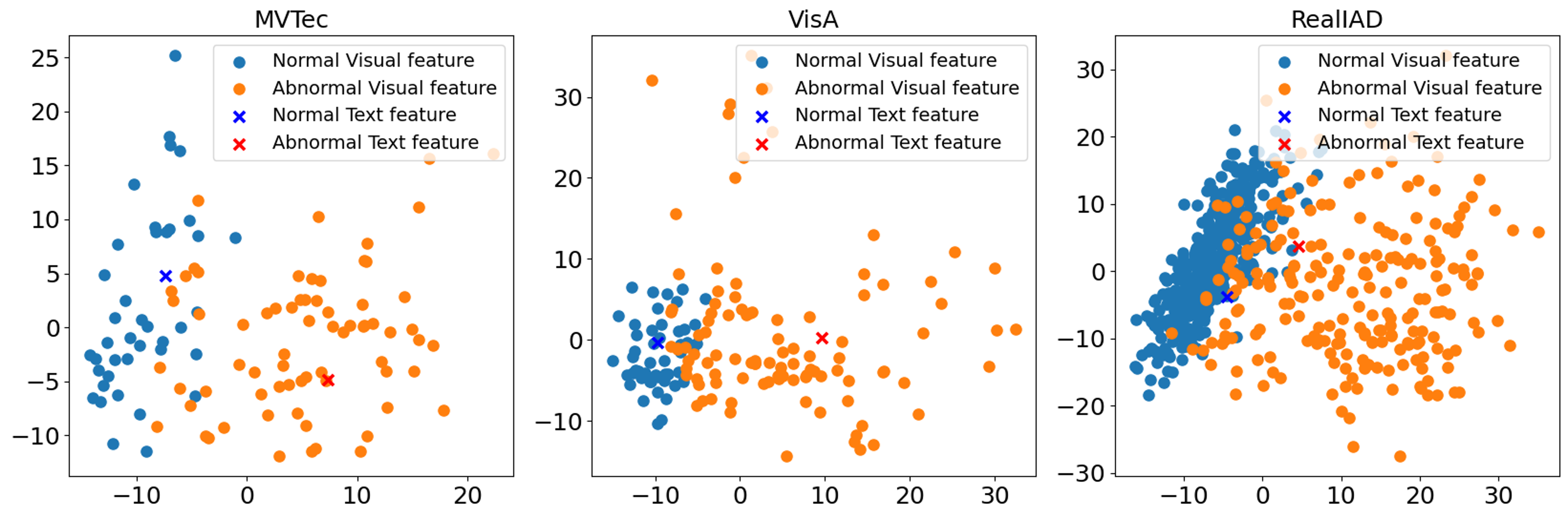}
    \caption{\textbf{Image-level Feature Visualization using PCA in a 1-shot setting.} The features ``hazelnut" for MVTec, ``pipe\_fryum" for VisA, and ``tape" for Real-IAD are used.}
    \label{fig:feat_vis}
\end{figure}

\begin{figure}
    \centering
    \includegraphics[width=\columnwidth]{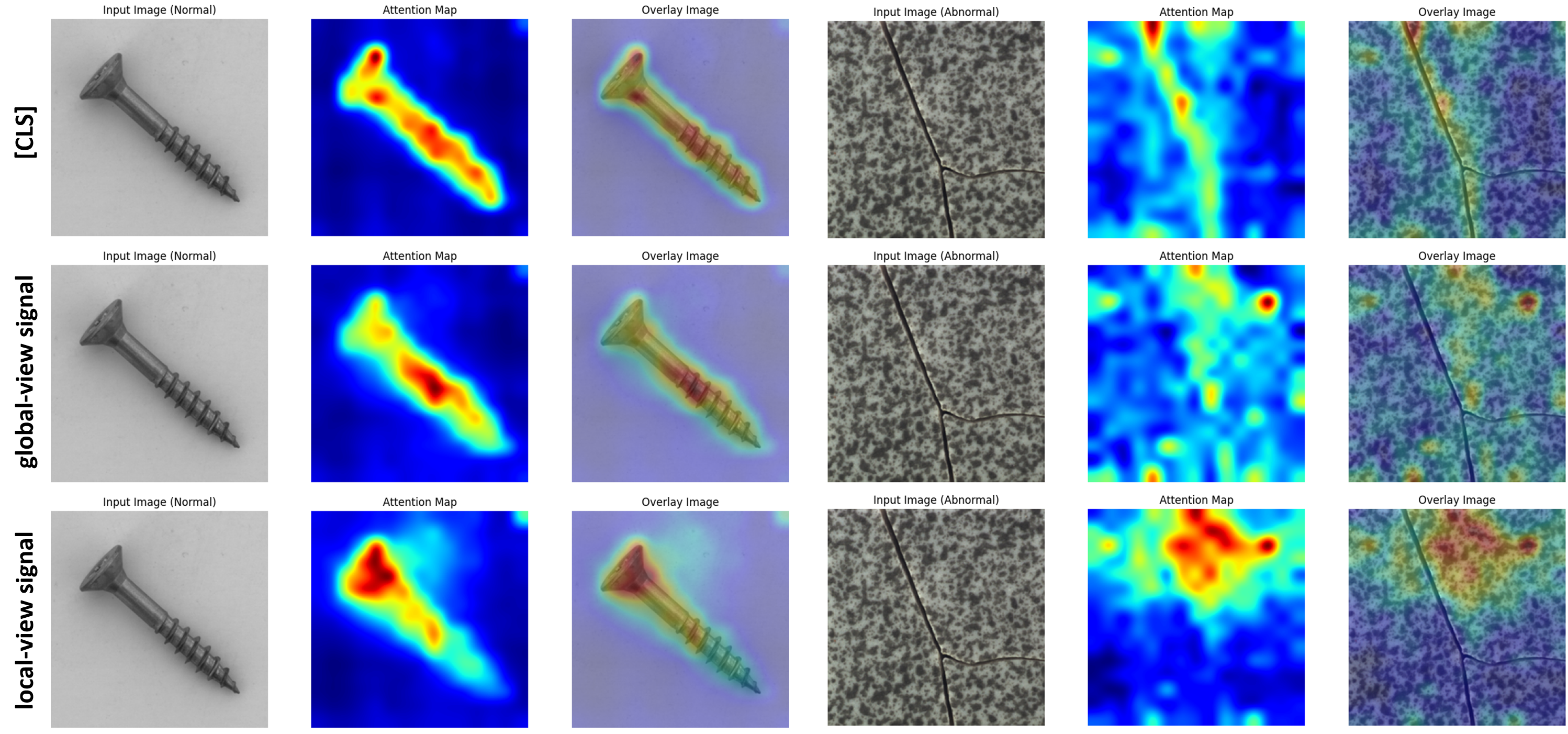}
    \caption{\textbf{Attention maps conditioned on the global-view prefix, local-view prefix, and the \texttt{[CLS]} token.} Columns 1, 4 are input images; columns 2, 5 display attention maps; columns 3, 6 show attention overlays.}
    \label{fig:multiview_signal}
\end{figure}

As shown in Figure \ref{fig:qualitative}, AnoPLe successfully localizes anomalous regions even in the extremely limited 1-shot setting, producing clean and spatially coherent heatmaps that closely align with the ground-truth regions. Despite the diverse texture, shape, and background variations across datasets, the model effectively suppresses false positives and captures fine-grained defect areas with high confidence. This demonstrates that the proposed multimodal prompt learning framework generalizes well across visual domains without category-specific tuning. Additional visual comparisons in the supplementary material show that AnoPLe provides consistently more accurate anomaly localization.

To confirm that our method performs well not only in pixel-level anomaly detection but also at the image level in a qualitative manner, Figure \ref{fig:feat_vis} visualizes image-level features using PCA in the 1-shot setting. The results show that AnoPLe produces well-separated clusters between normal and abnormal samples, and that textual features align naturally with their corresponding visual counterparts. This alignment confirms that the proposed bidirectional prompt learning not only enhances detection accuracy but also leads to semantically interpretable feature spaces across domains.

Beyond pixel- and image-level behaviors, we further analyze how AnoPLe utilizes multi-scale information during training. Since the model learns from both global images and local sub-images, but relies only on the global view during inference, it is essential to understand how these scale-dependent cues are organized internally. To this end, we visualize the attention maps conditioned on the learned scale-aware prefixes (Figure~\ref{fig:multiview_signal}). The global-view signal attends broadly to the main object, mirroring the holistic behavior expected at test time, while the local-view signal focuses on smaller patches consistent with its restricted spatial context. Notably, the \texttt{[CLS]} token, trained for anomaly detection, highlights anomalous regions in defective samples while maintaining object-level focus in normal ones. These visualizations confirm that AnoPLe successfully disentangles global and local cues during training, enabling stable inference despite the inherent train–test disparity in scale.

\subsection{Ablations}
\begin{table}[]
\centering
\caption{\textbf{Ablation on context vector interaction settings, multi-scale training, scale-aware prefix $c$, and prompt type.}}
\label{tab:ablations}
\resizebox{\linewidth}{!}{
\begin{tabular}{l|cc|cc|cc}
\toprule[1.5pt]
\multirow{2}{*}{\textbf{Setting}} & 
\multicolumn{2}{c|}{\textbf{MVTec}} &
\multicolumn{2}{c|}{\textbf{VisA}} &
\multicolumn{2}{c}{\textbf{RealiAD}} \\
& \textbf{I-AUC} & \textbf{P-AUC} & \textbf{I-AUC} & \textbf{P-AUC} & \textbf{I-AUC} & \textbf{P-AUC} \\
\midrule

\rowcolor{gray!15}
\multicolumn{7}{c}{\textbf{(a) Interaction Settings}} \\
Text & 93.0 & 95.0 & 81.7 & 95.7 & 79.9 & 96.5 \\
Image & 90.3 & {95.2} & 82.0 & 95.8 & {80.3} & 96.5 \\
T \& I (indep) & 92.4 & {95.2} & 80.3 & {96.0} & 80.1 & {96.6} \\
T → I & {93.5} & 95.1 & {84.2} & 95.9 & 80.1 & 96.5 \\
I → T & 91.2 & 95.0 & 82.0 & 95.9 & 80.0 & 96.5 \\
\textbf{T $\leftrightarrow$ I} & \textbf{94.5} & \textbf{96.0} & \textbf{86.0} & \textbf{97.0} & \textbf{81.2} & \textbf{96.7} \\

\hline
\rowcolor{gray!15}
\multicolumn{7}{c}{\textbf{(b) Alignment Strategy}} \\
No align & 93.7 & {95.3} & \textbf{86.0} & 95.9 & 73.3 & 94.9 \\
align with mean & {94.4} & 95.1 & {85.1} & {96.0} & {74.2} & {95.7} \\
\textbf{align with weighted sum} & \textbf{94.5} & \textbf{96.0} & \textbf{86.0} & \textbf{97.0} & \textbf{81.2} & \textbf{96.7} \\

\hline
\rowcolor{gray!15}
\multicolumn{7}{c}{\textbf{(c) Multi-scale Training / Prefix $c$}} \\
single-scale \& no prefix $c$& 89.9 & 95.1 & 82.3 & {95.8} & 79.5 & {96.5} \\
multi-scale \& no prefix $c$ & {91.8} & {95.2} & {83.3} & {95.8} & {80.0} & {96.5} \\
\textbf{multi-scale \& prefix $c$} & \textbf{94.5} & \textbf{96.0} & \textbf{86.0} & \textbf{97.0} & \textbf{81.2} & \textbf{96.7} \\

\hline
\rowcolor{gray!15}
\multicolumn{7}{c}{\textbf{(d) Prompt Type}} \\
\texttt{[object]} & 92.7 & 95.3 & 84.3 & 95.8 & 76.0 & 95.4 \\
\textbf{\texttt{[class]}}  & \textbf{94.5} & \textbf{96.0} & \textbf{86.0} & \textbf{97.0} & \textbf{81.2} & \textbf{96.7} \\
\bottomrule
\end{tabular}
}
\end{table}

Table \ref{tab:ablations}-(a) compares interaction strategies between textual and visual context vectors. Uni-modal prompts (``Text" or ``Image") perform consistently worse, whereas cross-modal interactions bring clear improvements. The bidirectional setting (T $\leftrightarrow$ I) achieves the best results across all benchmarks (94.5 / 96.0 on MVTec-AD, 86.0 / 97.0 on VisA, 81.2 / 96.7 on Real-IAD), showing that mutual refinement between modalities yields more discriminative anomaly representations. Independent or one-way variants provide partial gains but remain less effective. 

Table \ref{tab:ablations}-(b) evaluates how global features should be aligned with local evidence. Skipping alignment leads to weaker normality modeling, while mean alignment partially improves results. The weighted-sum alignment performs best on all datasets by incorporating anomaly-aware pixel prototypes, providing stronger guidance for refining global semantics. Table \ref{tab:ablations}-(c) shows that multi-scale training already boosts both detection and localization. Adding the scale-aware prefix $c$ yields the highest overall performance, indicating that explicit scale encoding helps maintain consistent anomaly cues across varying object sizes. 

Table \ref{tab:ablations}-(d) compares class-agnostic prompts (``object") with class-aware prompts (``class"). While the category-agnostic term appears to be more suitable for multi-class setting, class names consistently improve image-level performance (e.g., +5.2 I-AUROC on Real-IAD), as they serve as stronger semantic anchors for modeling normal object structure. We attribute this to the grounding effect of class names: they provide semantically richer anchors that help the visual prompts model normal object structure more precisely. This enhances sensitivity to subtle deviations while avoiding reliance on defect-specific anomaly descriptions, which remain harmful to generalization. Further discussion is provided in the supplementary material.

\begin{table}[]
    \centering
    \caption{\textbf{Comparison of anomaly detection and localization performance on three medical datasets in BMAD: Brain MRI, Liver CT, and Retinal OCT (4-shots).} The best score is in bold, and the second best is underlined.}
    \resizebox{\linewidth}{!}{
    \begin{tabular}{lcccccc}
        \toprule
        \multirow{2}{*}{\textbf{Method}} 
        & \multicolumn{2}{c}{\textbf{Brain MRI} \cite{baid2021rsna}} 
       & \multicolumn{2}{c}{\textbf{Liver CT} \cite{bilic2023liver,igelsias2015miccai}} 
       & \multicolumn{2}{c}{\textbf{Retinal OCT} \cite{hu2019automated}} \\
        \cmidrule(lr){2-3} \cmidrule(lr){4-5} \cmidrule(lr){6-7}
        & \textbf{I-AUC} & \textbf{P-AUC}& \textbf{I-AUC} & \textbf{P-AUC} & \textbf{I-AUOC} & \textbf{P-AUC} \\
        \midrule
        MedCLIP \cite{wang2022medclip} & 76.9 & 90.9 & 60.7 &  94.4 & 66.6 &  89.6 \\
         PatchCore \cite{roth2022towards} & 73.4 & 96.3 & 54.9 &  96.0 & 65.3 &  83.2 \\
         WinCLIP \cite{jeong2023winclip} & 53.7 & 86.2 & 61.9 &  97.8 & 75.4 &  94.4 \\
         PromptAD \cite{li2024promptadcvpr} & 73.9 & 94.3 & 62.1 &  96.4 & 91.2 &  96.6 \\
        UniVAD \cite{gu2025univad} & \textbf{82.6} & \underline{97.0} &  \underline{72.8} & \textbf{97.9} & \underline{88.9} & \underline{95.3} \\ \midrule
        \textbf{AnoPLe} & \underline{79.8} & \textbf{97.1} & \textbf{74.8 }& \underline{95.9} & \textbf{91.4} & \textbf{97.0}\\
        \bottomrule
    \end{tabular}
    }
    \label{tab:medical_bmad}
\end{table}

\subsection{Applicability to Medical Anomaly Detection}
Although AnoPLe is primarily designed for industrial domain, we demonstrate its applicability to the medical domain through the BMAD \cite{bao2024bmad} under a 4-shot setting (Table \ref{tab:medical_bmad}). Despite having access to only a few normal samples per class---and without incorporating any medical-specific adaptations---AnoPLe consistently delivers strong performance across Brain MRI, Liver CT, and Retinal OCT \cite{baid2021rsna,bilic2023liver,igelsias2015miccai,hu2019automated}, achieving the best or second-best results. Notably, AnoPLe outperforms both medical-specialized \cite{wang2022medclip} and industrial-domain–targeted anomaly detection approaches \cite{roth2022towards,jeong2023winclip,li2024promptadcvpr}, highlighting its robustness beyond its original design scope. Moreover, AnoPLe demonstrates competitiveness comparable to the latest universal anomaly detection framework UniVAD \cite{gu2025univad}, whose key strength is strong cross-domain generalization. These results indicate that AnoPLe is able to learn transferable normality cues, achieving robustness across visually diverse domains—even with limited data and no domain-specific tuning.

\section{Conclusion}
We present AnoPLe, a scalable framework for few-shot multi-class anomaly detection that removes the need for handcrafted anomaly descriptions through unified prompts and pseudo anomalies. By bidirectionally coupling textual and visual prompts, AnoPLe forms class-grounded and description-free representations that capture shared normal patterns across categories while detecting deviations purely from instance-level cues. To support fine-grained localization without additional inference cost, we incorporate a scale-aware prompting mechanism and a lightweight decoder that refines CLIP’s patch embeddings. Experiments on MVTec-AD, VisA, and Real-IAD show consistent gains over prior approaches in both detection and localization, along with strong robustness to unseen anomaly types, visually distinct domains, and even medical anomaly detection scenarios. By mitigating inter-class confusion and avoiding class-specific engineering, AnoPLe provides a simple, transferable, and practical solution for real-world industrial anomaly detection.
\section*{Acknowledgments}
This work was supported by the Institute of Information \& Communications Technology Planning \& Evaluation (IITP) grant funded by the Korean government (MSIT) (RS-2025-02305884).

{
    \small
    \bibliographystyle{ieeenat_fullname}
    \bibliography{main}

@String(AAAI = {AAAI})

@inproceedings{fang2023fastrecon,
  title={Fastrecon: Few-shot industrial anomaly detection via fast feature reconstruction},
  author={Fang, Zheng and Wang, Xiaoyang and Li, Haocheng and Liu, Jiejie and Hu, Qiugui and Xiao, Jimin},
  booktitle={Proceedings of the IEEE/CVF International Conference on Computer Vision},
  pages={17481--17490},
  year={2023}
}

@inproceedings{zhao2023omnial,
  title={Omnial: A unified cnn framework for unsupervised anomaly localization},
  author={Zhao, Ying},
  booktitle={Proceedings of the IEEE/CVF Conference on Computer Vision and Pattern Recognition},
  pages={3924--3933},
  year={2023}
}

@inproceedings{roth2022towards,
  title={Towards total recall in industrial anomaly detection},
  author={Roth, Karsten and Pemula, Latha and Zepeda, Joaquin and Sch{\"o}lkopf, Bernhard and Brox, Thomas and Gehler, Peter},
  booktitle={Proceedings of the IEEE/CVF Conference on Computer Vision and Pattern Recognition},
  pages={14318--14328},
  year={2022}
}

@article{cohen2020sub,
  title={Sub-image anomaly detection with deep pyramid correspondences},
  author={Cohen, Niv and Hoshen, Yedid},
  journal={arXiv preprint arXiv:2005.02357},
  year={2020}
}

@inproceedings{bergmann2019mvtec,
  title={MVTec AD--A comprehensive real-world dataset for unsupervised anomaly detection},
  author={Bergmann, Paul and Fauser, Michael and Sattlegger, David and Steger, Carsten},
  booktitle={Proceedings of the IEEE/CVF conference on computer vision and pattern recognition},
  pages={9592--9600},
  year={2019}
}

@inproceedings{zavrtanik2021draem,
  title={Draem-a discriminatively trained reconstruction embedding for surface anomaly detection},
  author={Zavrtanik, Vitjan and Kristan, Matej and Sko{\v{c}}aj, Danijel},
  booktitle={Proceedings of the IEEE/CVF International Conference on Computer Vision},
  pages={8330--8339},
  year={2021}
}

@inproceedings{gao2024learning,
  title={Learning to Detect Multi-class Anomalies with Just One Normal Image Prompt},
  author={Gao, Bin-Bin},
  booktitle={European Conference on Computer Vision},
  pages={454--470},
  year={2024},
  organization={Springer}
}

@inproceedings{rudolph2021same,
  title={Same same but differnet: Semi-supervised defect detection with normalizing flows},
  author={Rudolph, Marco and Wandt, Bastian and Rosenhahn, Bodo},
  booktitle={Proceedings of the IEEE/CVF winter conference on applications of computer vision},
  pages={1907--1916},
  year={2021}
}

@inproceedings{radford2021learning,
  title={Learning transferable visual models from natural language supervision},
  author={Radford, Alec and Kim, Jong Wook and Hallacy, Chris and Ramesh, Aditya and Goh, Gabriel and Agarwal, Sandhini and Sastry, Girish and Askell, Amanda and Mishkin, Pamela and Clark, Jack and others},
  booktitle={International conference on machine learning},
  pages={8748--8763},
  year={2021},
  organization={PMLR}
}

@inproceedings{jia2021scaling,
  title={Scaling up visual and vision-language representation learning with noisy text supervision},
  author={Jia, Chao and Yang, Yinfei and Xia, Ye and Chen, Yi-Ting and Parekh, Zarana and Pham, Hieu and Le, Quoc and Sung, Yun-Hsuan and Li, Zhen and Duerig, Tom},
  booktitle={International conference on machine learning},
  pages={4904--4916},
  year={2021},
  organization={PMLR}
}

@article{you2022unified,
  title={A unified model for multi-class anomaly detection},
  author={You, Zhiyuan and Cui, Lei and Shen, Yujun and Yang, Kai and Lu, Xin and Zheng, Yu and Le, Xinyi},
  journal={Advances in Neural Information Processing Systems},
  volume={35},
  pages={4571--4584},
  year={2022}
}

@inproceedings{huang2022registration,
  title={Registration based few-shot anomaly detection},
  author={Huang, Chaoqin and Guan, Haoyan and Jiang, Aofan and Zhang, Ya and Spratling, Michael and Wang, Yan-Feng},
  booktitle={European Conference on Computer Vision},
  pages={303--319},
  year={2022},
  organization={Springer}
}

@inproceedings{liu2023simplenet,
  title={Simplenet: A simple network for image anomaly detection and localization},
  author={Liu, Zhikang and Zhou, Yiming and Xu, Yuansheng and Wang, Zilei},
  booktitle={Proceedings of the IEEE/CVF Conference on Computer Vision and Pattern Recognition},
  pages={20402--20411},
  year={2023}
}

@inproceedings{
zhou2023anomalyclip,
title={Anomaly{CLIP}: Object-agnostic Prompt Learning for Zero-shot Anomaly Detection},
author={Qihang Zhou and Guansong Pang and Yu Tian and Shibo He and Jiming Chen},
booktitle={The Twelfth International Conference on Learning Representations},
year={2024},
url={https://openreview.net/forum?id=buC4E91xZE}
}

@inproceedings{gu2024anomalygpt,
  title={Anomalygpt: Detecting industrial anomalies using large vision-language models},
  author={Gu, Zhaopeng and Zhu, Bingke and Zhu, Guibo and Chen, Yingying and Tang, Ming and Wang, Jinqiao},
  booktitle={Proceedings of the AAAI Conference on Artificial Intelligence},
  volume={38},
  number={3},
  pages={1932--1940},
  year={2024}
}

@inproceedings{li2024promptadcvpr,
  title={Promptad: Learning prompts with only normal samples for few-shot anomaly detection},
  author={Li, Xiaofan and Zhang, Zhizhong and Tan, Xin and Chen, Chengwei and Qu, Yanyun and Xie, Yuan and Ma, Lizhuang},
  booktitle={Proceedings of the IEEE/CVF Conference on Computer Vision and Pattern Recognition},
  pages={16838--16848},
  year={2024}
}

@inproceedings{jeong2023winclip,
  title={Winclip: Zero-/few-shot anomaly classification and segmentation},
  author={Jeong, Jongheon and Zou, Yang and Kim, Taewan and Zhang, Dongqing and Ravichandran, Avinash and Dabeer, Onkar},
  booktitle={Proceedings of the IEEE/CVF Conference on Computer Vision and Pattern Recognition},
  pages={19606--19616},
  year={2023}
}

@inproceedings{zhu2024toward,
  title={Toward generalist anomaly detection via in-context residual learning with few-shot sample prompts},
  author={Zhu, Jiawen and Pang, Guansong},
  booktitle={Proceedings of the IEEE/CVF conference on computer vision and pattern recognition},
  pages={17826--17836},
  year={2024}
}

@article{deng2023anovl,
  title={Anovl: Adapting vision-language models for unified zero-shot anomaly localization},
  author={Deng, Hanqiu and Zhang, Zhaoxiang and Bao, Jinan and Li, Xingyu},
  journal={arXiv preprint arXiv:2308.15939},
  year={2023}
}

@inproceedings{zhou2022extract,
  title={Extract free dense labels from clip},
  author={Zhou, Chong and Loy, Chen Change and Dai, Bo},
  booktitle={European Conference on Computer Vision},
  pages={696--712},
  year={2022},
  organization={Springer}
}

@inproceedings{zou2022spot,
  title={Spot-the-difference self-supervised pre-training for anomaly detection and segmentation},
  author={Zou, Yang and Jeong, Jongheon and Pemula, Latha and Zhang, Dongqing and Dabeer, Onkar},
  booktitle={European conference on computer vision},
  pages={392--408},
  year={2022},
  organization={Springer}
}

@article{guo2024absolute,
  title={Absolute-Unified Multi-Class Anomaly Detection via Class-Agnostic Distribution Alignment},
  author={Guo, Jia and Han, Haonan and Lu, Shuai and Zhang, Weihang and Li, Huiqi},
  journal={arXiv preprint arXiv:2404.00724},
  year={2024}
}

@Article{Lv2025OneforAllFA,
 author = {Wenxi Lv and Qinliang Su and Wenchao Xu},
 booktitle = {International Conference on Learning Representations},
 title = {One-for-All Few-Shot Anomaly Detection via Instance-Induced Prompt Learning},
 year = {2025}
}

@inproceedings{he2024diffusion,
  title={A diffusion-based framework for multi-class anomaly detection},
  author={He, Haoyang and Zhang, Jiangning and Chen, Hongxu and Chen, Xuhai and Li, Zhishan and Chen, Xu and Wang, Yabiao and Wang, Chengjie and Xie, Lei},
  booktitle={Proceedings of the AAAI conference on artificial intelligence},
  volume={38},
  number={8},
  pages={8472--8480},
  year={2024}
}

@inproceedings{fan2025salvaging,
  title={Salvaging the Overlooked: Leveraging Class-Aware Contrastive Learning for Multi-Class Anomaly Detection},
  author={Fan, Lei and Huang, Junjie and Di, Donglin and Su, Anyang and Song, Tianyou and Pagnucco, Maurice and Song, Yang},
  booktitle={Proceedings of the IEEE/CVF International Conference on Computer Vision},
  pages={21419--21428},
  year={2025}
}

@inproceedings{wang2024real,
  title={Real-iad: A real-world multi-view dataset for benchmarking versatile industrial anomaly detection},
  author={Wang, Chengjie and Zhu, Wenbing and Gao, Bin-Bin and Gan, Zhenye and Zhang, Jiangning and Gu, Zhihao and Qian, Shuguang and Chen, Mingang and Ma, Lizhuang},
  booktitle={Proceedings of the IEEE/CVF Conference on Computer Vision and Pattern Recognition},
  pages={22883--22892},
  year={2024}
}

@inproceedings{wang2022medclip,
  title={MedCLIP: Contrastive Learning from Unpaired Medical Images and Text},
  author={Wang, Zifeng and Wu, Zhenbang and Agarwal, Dinesh and Sun, Jimeng},
  booktitle={Proceedings of the 2022 Conference on Empirical Methods in Natural Language Processing},
  pages={3876--3887},
  year={2022}
}

@inproceedings{gu2025univad,
  title={Univad: A training-free unified model for few-shot visual anomaly detection},
  author={Gu, Zhaopeng and Zhu, Bingke and Zhu, Guibo and Chen, Yingying and Tang, Ming and Wang, Jinqiao},
  booktitle={Proceedings of the Computer Vision and Pattern Recognition Conference},
  pages={15194--15203},
  year={2025}
}

@inproceedings{bao2024bmad,
  title={Bmad: Benchmarks for medical anomaly detection},
  author={Bao, Jinan and Sun, Hanshi and Deng, Hanqiu and He, Yinsheng and Zhang, Zhaoxiang and Li, Xingyu},
  booktitle={Proceedings of the IEEE/CVF Conference on Computer Vision and Pattern Recognition},
  pages={4042--4053},
  year={2024}
}

@article{baid2021rsna,
  title={The rsna-asnr-miccai brats 2021 benchmark on brain tumor segmentation and radiogenomic classification},
  author={Baid, Ujjwal and Ghodasara, Satyam and Mohan, Suyash and Bilello, Michel and Calabrese, Evan and Colak, Errol and Farahani, Keyvan and Kalpathy-Cramer, Jayashree and Kitamura, Felipe C and Pati, Sarthak and others},
  journal={arXiv preprint arXiv:2107.02314},
  year={2021}
}

@article{hu2019automated,
  title={Automated segmentation of macular edema in OCT using deep neural networks},
  author={Hu, Junjie and Chen, Yuanyuan and Yi, Zhang},
  journal={Medical image analysis},
  volume={55},
  pages={216--227},
  year={2019},
  publisher={Elsevier}
}

@article{bilic2023liver,
  title={The liver tumor segmentation benchmark (lits)},
  author={Bilic, Patrick and Christ, Patrick and Li, Hongwei Bran and Vorontsov, Eugene and Ben-Cohen, Avi and Kaissis, Georgios and Szeskin, Adi and Jacobs, Colin and Mamani, Gabriel Efrain Humpire and Chartrand, Gabriel and others},
  journal={Medical image analysis},
  volume={84},
  pages={102680},
  year={2023},
  publisher={Elsevier}
}

@inproceedings{igelsias2015miccai,
  title={Miccai multi-atlas labeling beyond the cranial vault--workshop and challenge},
  author={Igelsias, J and Styner, M and Langerak, T and Landman, B and Xu, Z and Klein, A},
  booktitle={Proc. MICCAI Multi-Atlas Labeling Beyond Cranial Vault—Workshop Challenge},
  year={2015}
}

@inproceedings{luo2025exploring,
  title={Exploring intrinsic normal prototypes within a single image for universal anomaly detection},
  author={Luo, Wei and Cao, Yunkang and Yao, Haiming and Zhang, Xiaotian and Lou, Jianan and Cheng, Yuqi and Shen, Weiming and Yu, Wenyong},
  booktitle={Proceedings of the Computer Vision and Pattern Recognition Conference},
  pages={9974--9983},
  year={2025}
}
}

\end{document}